\pdfoutput=1

\documentclass[11pt]{article}

\usepackage{acl}

\usepackage{times}
\usepackage{latexsym}

\usepackage[T1]{fontenc}

\usepackage[utf8]{inputenc}

\usepackage{microtype}

\usepackage{inconsolata}
\usepackage{graphicx}
\usepackage{amsmath}
\usepackage{booktabs}

\title{MapGuide: A Simple yet Effective Method to Reconstruct Continuous Language from Brain Activities}

\author{
Xinpei Zhao$^{1,2}$,
Jingyuan Sun$^{3}$\thanks{Corresponding Author}, 
Shaonan Wang$^{1,2}$\footnotemark[1],
Jing Ye$^{1,2}$,
Xiaohan Zhang$^{1,2}$, 
Chengqing Zong$^{1,2}$ \\
  \footnotesize${}^1$State Key Laboratory of Multimodal Artificial Intelligence Systems, Institute of Automation, CAS, Beijing, China\\
  \footnotesize${}^2$School of Artificial Intelligence, University of Chinese Academy of Sciences, Beijing, China\\
  \footnotesize${}^3$KU Leuven, Leuven, Belgium\\
  \footnotesize{\{zhaoxinpei2021, yejing2022\}@ia.ac.cn; jingyuan.sun@kuleuven.be} 
  \footnotesize{\{shaonan.wang, xiaohan.zhang, cqzong\}@nlpr.ia.ac.cn}
}

\begin{document}
\maketitle

\begin{abstract}

Decoding continuous language from brain activity is a formidable yet promising field of research. It is particularly significant for aiding people with speech disabilities to communicate through brain signals. This field addresses the complex task of mapping brain signals to text. The previous best attempt reverse-engineered this process in an indirect way: it began by learning to encode brain activity from text and then guided text generation by aligning with predicted brain responses. In contrast, we propose a simple yet effective method that guides text reconstruction by directly comparing them with the predicted text embeddings mapped from brain activities. Comprehensive experiments reveal that our method significantly outperforms the current state-of-the-art model, showing average improvements of 77\% and 54\% on BLEU and METEOR scores. We further validate the proposed modules through detailed ablation studies and case analyses and highlight a critical correlation: the more precisely we map brain activities to text embeddings, the better the text reconstruction results. Such insight can simplify the task of reconstructing language from brain activities for future work, emphasizing the importance of improving brain-to-text-embedding mapping techniques.

\end{abstract}

\section{Introduction}

Decoding continuous language text from brain activity stands as a groundbreaking endeavor at the nexus of neuroscience, linguistics, and artificial intelligence. 
Such an advancement promises to revolutionize communication, offering a new voice to those with speech impairments\cite{wolpaw_braincomputer_2002, haynes_decoding_2006}. 
Beyond enhancing communication, this research offers profound insights into the brain's language processing, paving the way for interfaces that integrate thought and speech effortlessly \cite{norman_beyond_2006, naselaris_encoding_2011, wang_computational_2024}.

While trials using invasive technologies like ECoG have shown promise\cite{willett_high-performance_2023}, the broad application of these methods is hampered by the limited public availability of invasive data and the complexities associated with neurosurgery. Decoding continuous language from non-invasive brain recordings, which are more accessible, remains a formidable challenge. This difficulty mainly stems from the intricate and dynamic relationship between language and the neural responses it elicits, further complicated by the inherently noisy nature of non-invasive neuroimaging.
The previous best attempt to tackle this issue first encoded brain activity from text with a linear model and then used this to guide text generation by aligning it with predicted brain responses \cite{tang_semantic_2022}. However, whether such an indirect method is optimal for the decoding task and whether a linear model is adequate for continuous text generation are questionable. Although this method has shown some improvement over random-level performance, the advancements are marginal.

Addressing the complex challenge of decoding continuous language from brain activities, we introduce \textit{\textbf{MapGuide}}, a simple yet effective two-stage framework. The first stage learns to \textit{\textbf{map}} brain activity to text embeddings with a Transformer-based mapper. We improve the mapper's resilience to neural noise by employing a random mask method for data augmentation and contrastive learning. In the second stage, a pre-trained text generator is \textit{\textbf{guided}} by text embeddings predicted with the mapper to produce text that closely aligns with the embeddings. MapGuide's integration of these two stages offers a more direct and effective solution for translating neural signals into a coherent text.

Experiments show that the proposed method achieves a new state-of-the-art (SOTA) result in reconstructing continuous language from fMRI-based brain recordings, significantly higher than the previous best attempt as measured by four different types of metrics. 
Our investigation further reveals an interesting contrast in compatibility patterns between frameworks: while previous encoding-based frameworks excel with linear models in linking brain activity and language, the decoding-based framework demonstrates superior performance when paired with non-linear models, underscoring a pivotal shift in approach for optimal results.
We also find a clear link between the accuracy of mapping brain activities to text embeddings and improved text reconstruction performance. This insight simplifies the task of reconstructing language from brain activities, emphasizing the importance of refining the brain-to-text embedding mapping process.

\section{Related Work}

\subsection{Reconstructing Language from fMRI}

The pioneering work of decoding language from fMRI-recorded brain activities can be traced back to Michael et al.'s paper in  \citeyear{mitchell_predicting_2008}. Since then, this area has primarily focused on word-level and single-sentence-level decoding, greatly enhancing our understanding of neural representations. Initially, fMRI decoding at the word level was approached through pairwise classification, choosing the most appropriate word from a pair \cite{mitchell_predicting_2008, palatucci_zero-shot_2009}. Some work comprehensively explained the influence of different factors on word decoding \cite{wang_fine-grained_2020}.
More recent efforts have concentrated on aligning cognitive signals with a limited vocabulary, typically up to a thousand words for word-level decoding \cite{defossez_decoding_2023}, or incorporating these into sentence embeddings for sentence-level decoding, also using pairwise classification \cite{pereira_toward_2018, sun_towards_2019, sun_neural_2021}. 
The latest research in this field has been exploring various strategies for decoding fMRI to text, including prompt-based and direct decoding approaches \cite{zou_towards_2021, zou_cross-modal_2022, tang_semantic_2022, xi-etal-2023-unicorn}.

\subsection{Text Generation with Pre-trained Language Model}

The field of neural decoding has significantly advanced with the emergence of pre-trained language models. Generative models like GPT \cite{radford_improving_2018} and GPT2 \cite{radford_language_2019} have become especially notable for their capacity to produce coherent, contextually relevant text, aligning closely with underlying neural patterns. Additionally, applying BART \cite{lewis_bart_2020} in fMRI decoding has proven its effectiveness in generative decoding tasks. This further emphasizes the crucial role of pre-trained language models in progressing the realm of fMRI decoding.

\section{Methodology}

\begin{figure*}[t]
    \centering
    \includegraphics[width=1.05\textwidth]{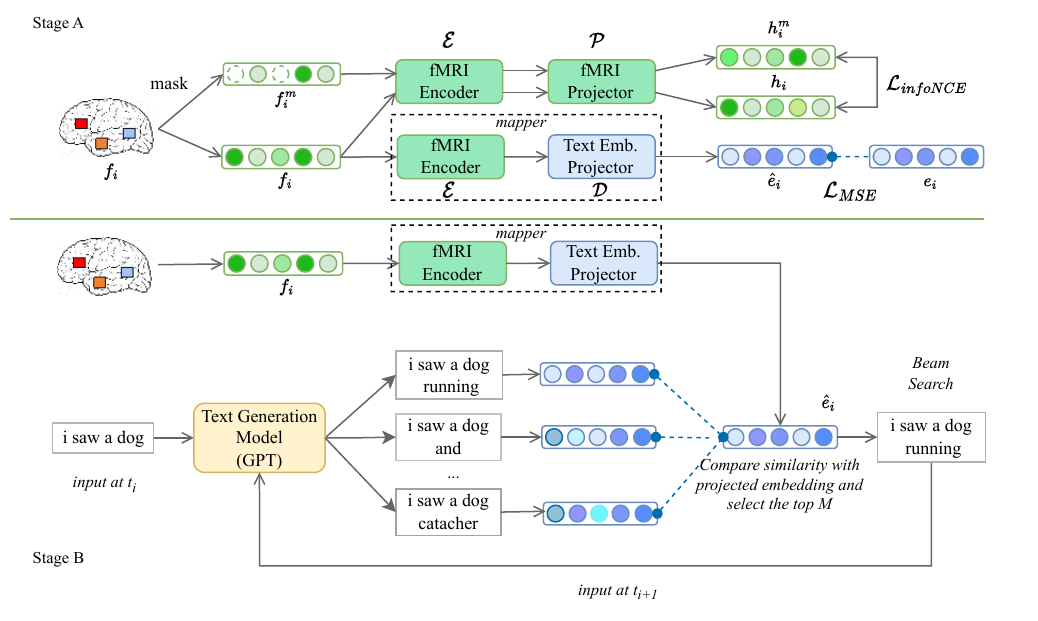} 
    \caption{Structure of MapGuide to generate text from brain imaging. Stage A maps brain imaging to text embeddings, while stage B generates texts under the guidance of the mapper.}
    \label{fig: main structure of MapGuide}
\end{figure*}

\subsection{Motivation and Overview}
\label{sec: motivation}

In this section, we explore the characteristics of fMRI data and brain language representations that have led to the development of our MapGuide framework.

Firstly, the relationship between input language and the aroused neural responses is non-linear, highly complex, and dynamic. Secondly, fMRI recordings are characterized by their inherent noise, arising from various physiological and scanner-related sources. While these recordings capture responses to linguistic stimuli, they also pick up signals from various other cognitive activities. Last but not least, fMRI primarily measures changes in blood-oxygen-level-dependent (BOLD) signals. A common observation in fMRI data is the similarity in signal magnitudes across adjacent voxels, indicating a level of spatial redundancy.

To address these challenges, we introduce MapGuide, a direct two-stage framework as illustrated in Figure \ref{fig: main structure of MapGuide}. Stage A employs a Transformer-based mapper to map brain activity to text embeddings, aiming to capture the complex brain-language interaction. We further apply contrastive learning with random masking, targeting the noises and spatial redundancy of fMRI data. Stage B then guides text generation using a pre-trained text generator, guided by the learner mappers in Stage A, making full use of recent developments of large language models.

\subsection{Stage A: Mapping from Brain Activities to Text Embeddings}
We have developed a mapper to predict fMRI to text features (see Appendix \ref{app: text features} for details) while enhancing fMRI representation robustness. Our mapper consists of the fMRI encoder \(\mathcal{E}\) and the text embedding projector \(\mathcal{D}\). The encoder \(\mathcal{E}\) processes fMRI data into a latent space, which the text embedding projector then translates into text embeddings. 
We first optimize the Mean Squared Error (MSE) loss between predicted and ground truth text embeddings. 
The loss \(\mathcal{L}_{MSE}\) is formulated as:
\begin{equation}
    \hat{T} = \mathcal{D}( \mathcal{E}(F) ) \\
\end{equation}
\begin{equation}
    \mathcal{L}_{MSE} = \frac{1}{N} \sum_{i=1}^{N} \sum_{j=1}^{D} (e_{ij} - \hat{e}_{ij})^2 \\
\end{equation}
where $N$ and $D$ denote the batch size and dimension number of text embedding while $e$ and $\hat{e}$ are the respectively the ground truth and predicted embeddings.

To learn denoised fMRI representations, our model further incorporates contrastive learning with random masking technique \( M(\cdot, \text{ratio}) \), generating masked data \(F^{m} = M(F, \text{ratio})\) and treating masked samples as positive samples to the unmasked input. 
Without loss of generality, we use infoNCE loss as the loss function for contrastive learning\cite{oord_representation_2019}.
The fMRI projector \(\mathcal{P}\) derives the hidden layer representation, with the infoNCE loss calculated as follows:
\begin{equation}
    H = \mathcal{P}(\mathcal{E}(F)) 
\end{equation}
\begin{equation}
    H^{m} = \mathcal{P}(\mathcal{E}(F^{m})) 
\end{equation}
\begin{equation}
    \mathcal{L}_{infoNCE} = -\sum_{i=1}^{N} \log \frac{\exp (h_i \cdot h_i^{m} / \eta)}{\sum_{j=1}^{N} \exp (h_i \cdot h_j^{m} / \eta)}
\end{equation}
where \(N\) is the number of samples in the batch, and \(\eta\) is the temperature parameter for the InfoNCE loss.

A hybrid loss function combining \(\mathcal{L}_{MSE}\) and \(\mathcal{L}_{infoNCE}\) ensures accurate text reconstruction and effective differentiation between samples.

\subsection{Stage B: Guiding Language Generation with the Mapper}

In Stage B, following the acquisition of text representations in Stage A, we use a pre-trained generative language model for text generation, as illustrated in Figure \ref{fig: main structure of MapGuide}. The model, implementing a beam search algorithm \cite{tillmann_word_2003}, generates multiple continuations for each sequence in the beam at each time step. We then evaluate the similarity of these continuations to our predicted text representation, retaining the most likely ones for the next step. This process iteratively continues, aligning the generated text with the brain's representations until the sequence is complete.

\section{Experimental Setup}

In this section, we will first introduce the task and the fMRI dataset, then describe evaluation metrics, baselines to be compared, and implementation details.

\subsection{Task}

The text decoding task involves analyzing a series of fMRI images paired with corresponding timestamps to reconstruct the text heard by a subject at specific times. This process is represented as $\mathcal{F}:=\{(f_1, \tau_1), (f_2, \tau_2), \ldots, (f_m, \tau_m) \}$, where each pair $(f_i, \tau_i)$ corresponds to an fMRI image taken at time $\tau_i$, with $m$ being the total number of images. The aim is to predict a series of words and their timings, denoted as $\mathcal{W}:=\{(w_1, t_1), (w_2, t_2), \ldots, (w_n, t_n) \}$, where each $(w_i, t_i)$ indicates the predicted word $w_i$ at time $t_i$, and $n$ is the number of words in the prediction sequence.

The fMRI images are captured at consistent time intervals, known as the repetition intervals (TR), ensuring uniformity in $\{ \tau_i \}$. Predicting each word's timing in $\mathcal{W}$ uses a linear word rate model, focusing on the brain's auditory cortex to maintain uniformity in $\{ t_i \}$. The task's goal is to identify words and their timings that closely resemble the original word series, mathematically formulated as:

\begin{equation}
   \hat{w_i} = \underset {w_i} {\mathrm{argmax}} \, p(w_i | \mathcal{F}, t_i)
\end{equation}

\begin{table*}[t]
    \centering
    \includegraphics[width=\textwidth]{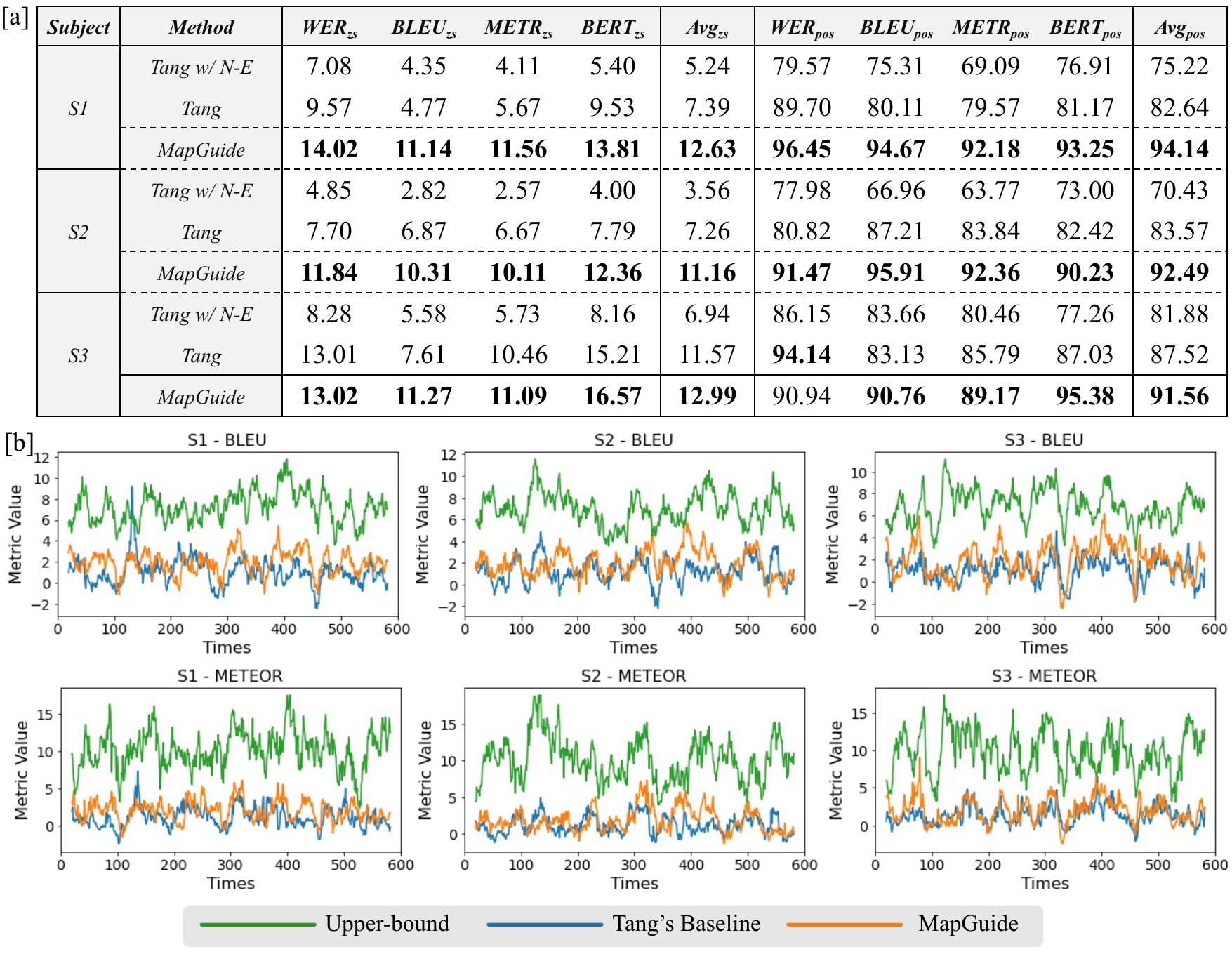} 
    \caption{
    Evaluation metrics of decoding results. [a] Word-error rate (WER), BLEU, METEOR (METR), and Bert-score (BERT) of decoding results. w/N-E denotes with non-linear encoder. Definitions of metrics are detailed in Section 4.3. [b] BLEU and METEOR (Story-Zscore) of the optimal upper-bound, Tang's baseline, and our MapGuide as depicted along window times. 
    }
    \label{tab: main results}
\end{table*}

\subsection{Dataset}

We use the dataset provided by \citet{lebel_natural_2023} to evaluate decoding performance, concentrating on perceptual speech task responses. 
The dataset includes a set of training stories and one test story, comprising 27,449 fMRI samples in the training set and 291 in the test set, collected from three subjects who listened to identical training and testing stories. 
The fMRI data were acquired with a TR of 2 seconds. 
The dataset also includes word information and timestamps for each word. 
We select 10,000 cortical voxels for decoding purposes, consistent with those chosen by \citeauthor{tang_semantic_2022}.

\subsection{Metrics}
\label{Sec: Metrics}

Text generation quality is assessed by comparing generated text to the actual text using a 20-second sliding window. The word order in each window is categorized into reference (ground truth) and prediction columns. To establish a baseline, 200 random sequences are generated, and their average performance is used for comparison.

We calculate two primary metrics: the positive rate and the story-zscore. The positive rate measures how frequently the similarity between the reference and prediction exceeds a certain threshold, indicating minor (micro) improvements in decoding performance over random. The story-zscore assesses the overall (macro) improvement by calculating the deviation of predicted similarity from the average similarity of all random sequences. To evaluate the similarity of reference and prediction, various language similarity metrics are employed, including Word Error Rate (WER), BLEU-1 (BLEU)\cite{papineni_bleu_2002}, METEOR (METR)\cite{banerjee_meteor_2005}, and BERTScore (BERT)\cite{zhang_bertscore_2020}. For more details, see the Appendix \ref{app: details of metrics}. 

\begin{table*}[t]
    \centering
    \includegraphics[width=6.3 in]{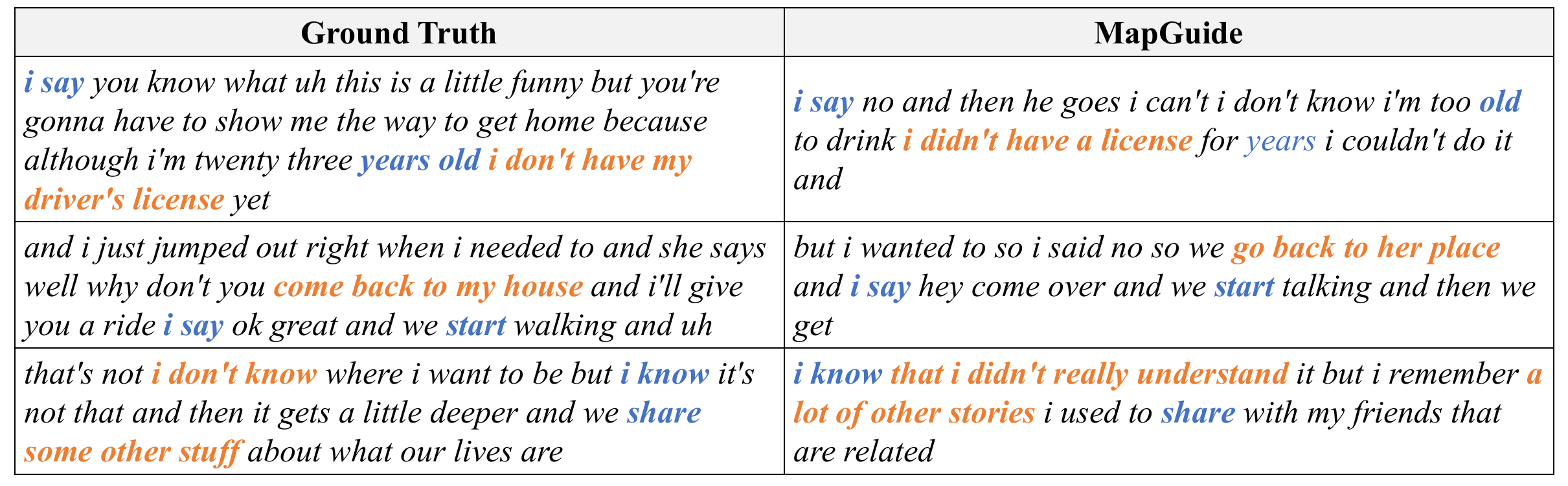} 
    \caption{
    Samples of reconstructed language text. 
    The highlighted text in blue and orange represents the parts of the MapGuide that are semantically identical or similar to the ground-truth text.
    }
    \label{tab: demo of decoding}
\end{table*}

\begin{table*}[t]
    \centering
    \includegraphics[width=6in]{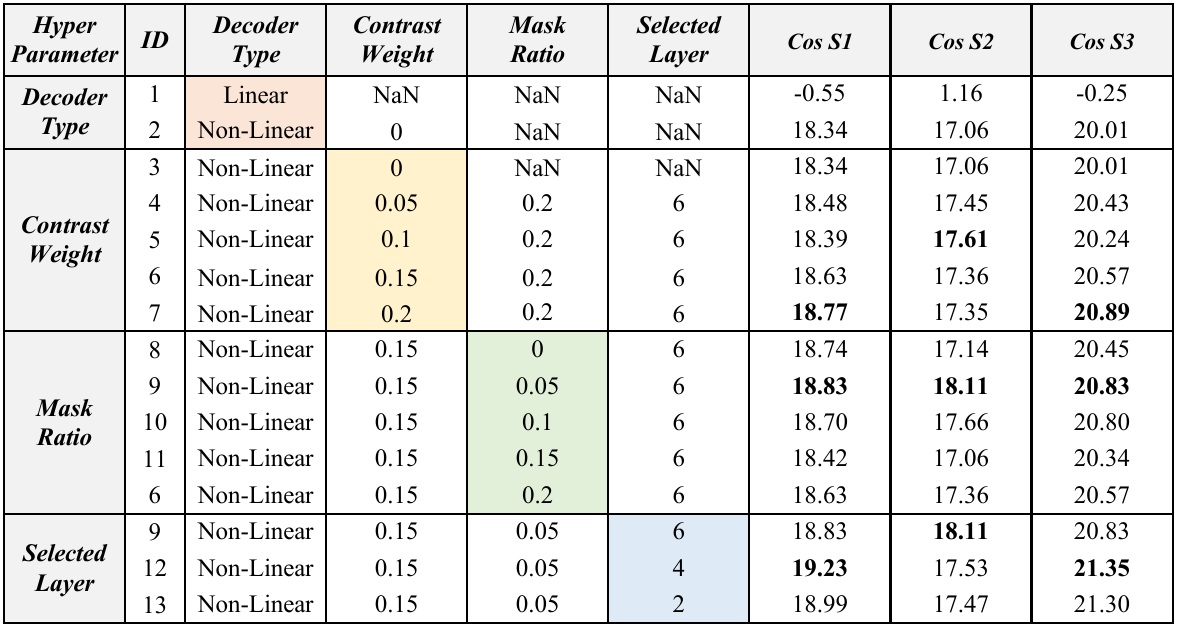} 
    \caption{Results of ablation experiments in Stage A on all three subjects S1-S3. We use the cosine similarity between predicted and ground truth text embeddings as the metric. Cells with colored shades denote the hyper-parameters tuned in one ablation group and resulting metrics. For example, cells with green shades denote that mask ratio is the parameter to be tuned while other parameters are kept the same.}
    \label{tab: ablation study1}
\end{table*}

\begin{table*}[t]
    \centering
    \includegraphics[width=5.9in]{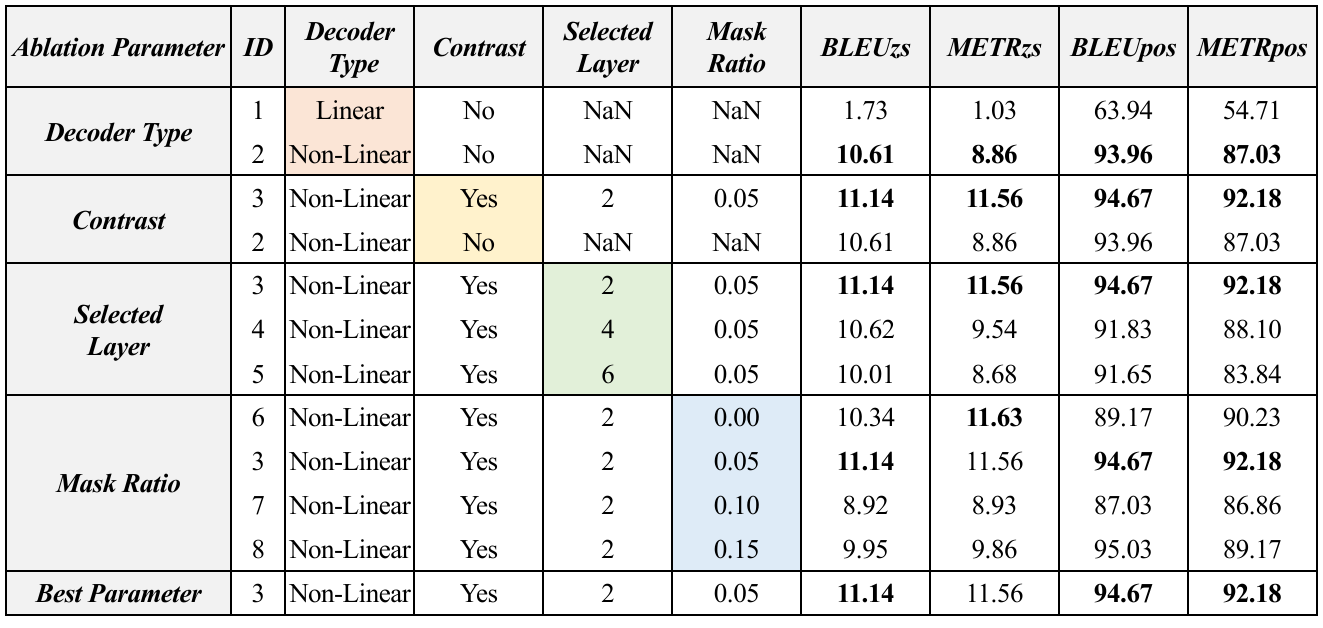} 
    \caption{Results of ablation experiments for text reconstruction. We use BLEU and METR for the paramount accuracy of the ablation. Cells with colored shades denote the hyper-parameters tuned in one ablation group and resulting metrics. For example, cells with green shades denote that mask ratio is the parameter to be tuned while other parameters are kept the same.}
    \label{tab: ablation study2}
\end{table*}

\subsection{Baseline}

In our study, we have chosen the approach by \citeauthor{tang_semantic_2022} as the baseline for text decoding. This decision was influenced by two key factors. Firstly, our decoding and evaluation process is uniquely designed to be stepwise and timestamp-based. This choice is based on the practical application scenarios of neural decoding, particularly in the field of brain-computer interfaces. For practical human needs, the ability to collect signals step-by-step during human language expression and generate words accordingly is more in line with the application scenario of human conversation. Therefore, starting from practical application scenarios, we prefer to choose a task form that is closer to real-world use. Secondly, the multi-subject model utilized in studies such as \citet{xi-etal-2023-unicorn} significantly differs from the single-subject focus of our task, making them less suitable for direct comparison.

\subsection{Implementation}

To implement Stage A, we employ the Huggingface Transformers library based on PyTorch.
While training the models in Stage A, we partition the datasets into training and validation sets with an 80\% - 20\% split ratio. 
We train separate models for each of the three subjects. 
For Stage B, we use the same structure of our baseline. Since changing the hyperparameters of Stage B will affect the random generation, we use the same hyperparameters by Tang's method. 
Additionally, we leverage the pre-trained GPT model introduced by Tang for two purposes: feature extraction and text generation.
All experimental procedures are carried out using eight NVIDIA GeForce GTX 1080 Ti GPUs.
The hyper-parameter settings to achieve the best performance are detailed in the ablation study in Section \ref{sec: ablation study}.

\begin{figure*}[t]
    \centering
    \includegraphics[width=6.3 in]{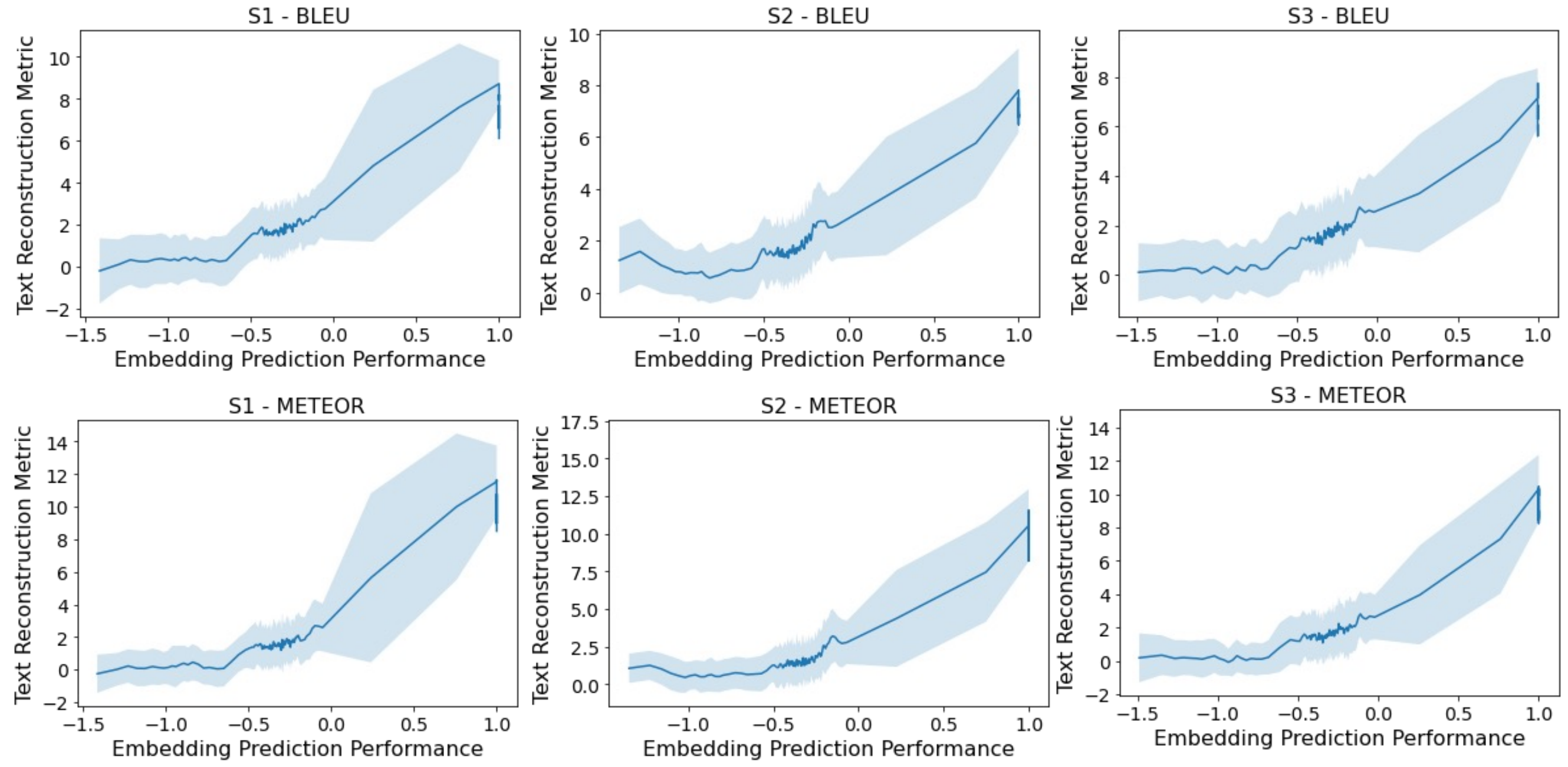} 
    \caption{
    An error band line chart depicting the relationship between the embedding prediction performance of Stage A's learned mapper and the final text reconstruction metrics. We plot with the BERT and METEOR scores on all three subjects, with the statistical scope covering all models mentioned in the ablation experiments and the optimal upper-bound. The values of embedding prediction performance have been standardized.
    }
    \label{fig: correlation}
\end{figure*}

\section{Results}

In this section, we will first compare the metrics of text reconstruction of MapGuide against the previous SOTA. 
We will then conduct a detailed ablation study to evaluate the efficacy of MapGuide's modules and discuss the influence of hyper-parameter setting. 
We will lastly do a correlation analysis of the experimental results. 

\subsection{Reconstruction Results}

Experiment results of metrics are shown in Table \ref{tab: main results}[a]. The results show that the performance of the nonlinear-based decoding model is significantly better than those of the other two models.  
Our model achieves an accuracy exceeding that of Tang's method in the story-zscore of BLEU by \(77\%\) (calculated by \((11.14-4.77)/4.77 + (10.31-6.87)/6.87 + (11.27-7.61)/7.61 \approx  77\%\) ) and of METEOR by \(54\%\).

The BLEU and METEOR scores along window times are shown in Table \ref{tab: main results}[b].
As can be seen from the line chart, our method significantly outperforms the baseline model at many time points in all three subjects. The advantages of the non-linear decoding model framework are verified through experiments. 
Meanwhile, the superior performance of the upper bound indicates the great potential of the neural decoding-based framework. 
The experimental results consistently show that the non-linear decoding model is superior in most cases. Interestingly, linear encoding models rank better than non-linear encoding models, consistent with the hypothesis that non-linear models are better suited for high-to-low-dimensional tasks. In contrast, linear models excel in low-to-high-dimensional tasks.

To have a qualitative impression of the text reconstructed by our model, we randomly select some samples and depict them in Table \ref{tab: demo of decoding}. As shown in the samples, though not fully resembling the ground truth, the texts reconstructed by our model have several fragments semantically identical or similar to the ground truth.  

\subsection{Ablation Study}
\label{sec: ablation study}
Our framework is a two-stage pipeline. In this section, we will conduct an ablation study to specify the effects of hyper-parameter settings on Stage A's intermediate results and Stage B's final text reconstruction performance.

\subsubsection{Effects of Hyper-paramters on Intermediate Results}

In Stage A, we train a Transformer-based mapper to predict text embeddings from brain activities with contrastive learning. This stage has three essential hyper-parameters: the weight of contrastive loss, the masking ratio for contrastive learning, and the layer of the fMRI encoder connected to the fMRI projector. We will study the effects of tuning these hyper-parameters according to how they influence the quality of mapped text embeddings. We use the cosine similarity between the mapped and ground truth text embeddings as the metric to assess these intermediate results.

\noindent\textbf{Effects of Tuning Contrastive Loss Weight}

The weight of contrastive loss conditions the importance of learning denoised fMRI representations in Stage A. As shown in Table \ref{tab: ablation study1} experiments 3 - 7, setting the largest weight of 0.2 yields the best text embeddings on two of the three subjects. On only subject 2, a medium weight of 0.1 yields better performance. On all the subjects, setting the weight as 0 predicts the worst quality embeddings. These results demonstrate the importance of using contrastive learning to produce denoise representations.

\noindent\textbf{Effects of Tuning Masking Ratio}

In Stage A, we are conducting contrastive learning with masked fMRI; the masking ratio on fMRI is thus a critical hyper-parameter to be considered. As shown in Table \ref{tab: ablation study1} experiments 6,8,9,10,11, a mask ratio of 5\% has been enough to achieve the best embedding mapping performance on all subjects. Not using masking or a more considerable masking tends to degrade the performance. This is within expectation. As we introduce in Section \ref{sec: motivation}, masking may help target the spatial redundancy in fMRI. However, masking too much on the neuro-image could cause a loss of information and introduce further noise. 

\noindent\textbf{Effects of Layer Selection}

As shown in Figure \ref{fig: main structure of MapGuide}, by default, the output of the fMRI encoder will be input to the fMRI projector to conduct contrastive learning. However, we are curious if using the output shallower layers of the fMRI encoder could yield better performance since we may learn denoised fMRI representations at earlier stages. So in Table \ref{tab: ablation study1} experiments 9,12 and 13, we select different layers of fMRI encoder of which output is fed to the fMRI projector. We find that on two subjects, selecting a medium layer leads to the best embedding predictions.

\subsubsection{Effects of Hyper-parameters on Final Text Reconstruction}

To ensure a fair comparison with the previous SOTA approach, we follow their settings of hyper-parameters for the text generation model in Stage B. So, the hyper-parameters that will largely influence the text reconstruction performance are for Stage A's mapper model. We use the mappers trained with different sets of hyper-parameters to guide the text generation and present the performance of Stage B in Table \ref{tab: ablation study2}. Due to space limits, we only present the results on subject S1 without losing generalizability.

We found that the hyper-parameters yielding the best intermediate results in Stage A still mostly lead to better text generation performance, and vice versa. For example, in Table \ref{tab: ablation study2}'s experiments 3,6,7,8 that display the effects of mask ratio, we still find that a mask ratio of 5\% yields the best final text reconstruction performance. Replacing our proposed mapper with a linear regression model yields the worst embedding prediction performance, as shown in Table \ref{tab: ablation study1}'s experiment 1. It also leads to the lowest text reconstruction accuracy in Stage B, as depicted in. However, there are also minor exceptions. In Table \ref{tab: ablation study1}, we find that selecting medium layers of fMRI encoder for contrastive learning leads to better embedding predictions. However, in Table \ref{tab: ablation study2}'s experiments 3-5, introducing contrastive learning even earlier leads to better final text reconstruction accuracy. 
In the following case study section, we will check if there were correlations between the embedding prediction performance in Stage A and the final text reconstruction performance in Stage B.

\subsection{Correlation Analysis}

In previous sections, we observe a tendency that the better the mapper performs in Stage A to predict text embeddings, the more likely the mapper can guide a better text reconstruction in Stage B. This is also intuitive since a high-quality mapper could more accurately guide the text generator to reconstruct semantic-related contents. In this section, we will check whether such intuitions comply with our experimental results.

We present an error band line chart in Figure \ref{fig: correlation}, using the normalized embedding prediction performance of the mapper in Stage A as the X-axis and the final text reconstruction performance in Stage B as the Y-axis. Like in prior sections, we still use the cosine similarity of predicted and ground-truth embeddings to measure the performance of the mapper. We plot the line charts with the experimental results of all three subjects. The blue lines in Figure \ref{fig: correlation} fit our real experiment results, while the shades reflect the variance. Figure \ref{fig: correlation} shows a clear positive correlation between the embedding prediction performance and text reconstruction metric. This is an informative finding of our work. Following this finding, we can simplify the highly complex task of decoding continuous text from brain activities by focusing on improving the mapping from neural activations to text embeddings.

\section{Conclusion}

In this paper, we propose MapGuide, a simple yet effective double-stage framework for reconstructing continuous language from brain activities. In the first stage, we learn a mapper that decodes text embeddings from brain activities with contrastive learning. In the second stage, the mapper is applied to supervise a text generation model. MapGuide exceeds the previous SOTA by a large margin on all evaluation metrics. Through comprehensive ablation studies and in-depth case analyses, we further substantiate the efficacy of MapGuide's modules. Our research further reveals a direct correlation between the precision of mapping brain activities to text embeddings and the subsequent improvements in text reconstruction performance. This insight can be informative in streamlining the intricate process of language reconstruction from brain activities. By enhancing the mapping from brain activities to text embeddings, we can significantly simplify and improve the task of language reconstruction.

\section*{Limitation}

To date, our testing has been limited to English single-subject datasets. Expanding our analysis to encompass single-subject data in languages other than English, such as Chinese, presents a promising avenue for future research\cite{wang_synchronized_2022}. Nevertheless, our current approach has yet to undergo validation from a cross-lingual perspective.

Additionally, we have yet to explore utilizing more intricate structures for fMRI reconstruction extensively. Previous research has demonstrated the efficacy of pre-training-based architectures in image decoding \cite{chen_cinematic_2023, sun_contrast_2023, sun_decoding_2023, sun_neurocine_2024}. In our forthcoming work, we explore incorporating more complex reconstruction methods.

\section*{Acknowledgements}
We would like to thank the anonymous reviewers for their valuable comments. 

This research was supported by grants from the National Natural Science Foundation of China to S. W. (62036001) and   S.W. (the STI2030-Major Project, grant number: 2021ZD0204105).

\bibliography{anthology, custom}

\appendix
\section{Details of Metrics}
\label{app: details of metrics}

We assess text generation quality by comparing the generated text's word time series with the actual text, following the methodology proposed by \citeauthor{tang_semantic_2022}. This evaluation employs a fixed-length window (20 seconds) that slides along both the ground truth sequence $\mathcal{W}$ and the predicted sequence $\hat{\mathcal{W}}$, covering the period from $t_{\text{start}}$ to $t_{\text{end}}$. In each window, the word order is categorized into two columns: $R_i$ for the reference (ground truth) sequence and $P_i$ for the prediction. 
To establish a baseline for comparison, we generate 200 random sequences. The average performance of these sequences serves as the benchmark for random performance.

The formulas for these metrics are:
\begin{equation}
\text{sim}_{\text{pos}} = \frac{ |\{ R_i \in R: \text{sim}(R_i, P_i) - \mu_i > 0\}|}{S}
\end{equation}
\begin{equation}
\text{sim}_{\text{ZS}} = \frac{\frac{\sum_i{\text{sim}(R_i, P_i)}}{S} - \mu }{\sigma}
\end{equation}
In these equations, $S$ is the total number of windows, $\text{sim}(R_i, P_i)$ represents the similarity between $R_i$ and $P_i$, and $\mu_i$ and $\sigma_i$ are the mean and standard deviation of similarity for each window, respectively. $\mu$ and $\sigma$ denote the average similarity's mean and standard deviation across all random sequences.

Various language similarity metrics are employed in the $\text{sim}(R_i, P_i)$ measure, including Word Error Rate (WER), BLEU-1 (BLEU)\cite{papineni_bleu_2002}, METEOR (METR)\cite{banerjee_meteor_2005}, and BERTScore (BERT)\cite{zhang_bertscore_2020}. 
WER calculates the number of edit operations needed to transform the prediction into the reference. BLEU counts the occurrences of predicted unigrams in the reference, measuring precision. METEOR considers synonyms and stemming, combining predicted and reference unigrams. BERTScore uses contextualized embeddings for recall, applying inverse document frequency (IDF) importance weighting computed across the training dataset's stories.

\section{Acquisition of Text Embeddings}
\label{app: text features}

We replicate the methodology outlined in prior research by \citeauthor{tang_semantic_2022} to generate a stimulus matrix that corresponds to the fMRI data. For each word-time pair $(s_i, t_i)$ within every narrative, we input the word sequence $(s_{i-5}, s_{i-4}, \ldots, s_{i-1}, s_i)$ into a language model. From the model's hidden layer, we extract semantic features of $s_i$, resulting in a revised list of vector-time pairs $(M_i, t_i)$, where $M_i$ signifies an $n$-dimensional semantic embedding for $s_i$. These pairs are resampled utilizing a three-lobe Lanczos filter to synchronize the vectors with the fMRI acquisitions. Subsequently, we employ a linearized Finite Impulse Response (FIR) model to fit every cortical voxel in each subject's brain\cite{huth_natural_2016}. For each of the $n$ features, we apply a distinct linear temporal filter with four delays ($t - 1$, $t - 2$, $t - 3$, and $t - 4$ timepoints), resulting in a total of $4n$ features. All punctuation is removed during the representation acquisition and text generation process. 

\end{document}